\def\year{2021}\relax
\documentclass[letterpaper]{article} % DO NOT CHANGE THIS
\usepackage{aaai21}  % DO NOT CHANGE THIS
\usepackage{times}  % DO NOT CHANGE THIS
\usepackage{helvet} % DO NOT CHANGE THIS
\usepackage{courier}  % DO NOT CHANGE THIS
\usepackage[hyphens]{url}  % DO NOT CHANGE THIS
\usepackage{graphicx} % DO NOT CHANGE THIS
\usepackage{natbib}  % DO NOT CHANGE THIS AND DO NOT ADD ANY OPTIONS TO IT
\usepackage{caption} % DO NOT CHANGE THIS AND DO NOT ADD ANY OPTIONS TO IT
\usepackage{times}  %Required
\usepackage{helvet}  %Required
\usepackage{courier}  %Required
\usepackage{url}  %Required
\usepackage{graphicx}  %Required
\usepackage{lipsum}
\usepackage{amsmath,amssymb,amsthm}
\usepackage{natbib}
\usepackage{color}
\usepackage[]{algorithm2e}
\usepackage{nicefrac}
\usepackage{caption}
\usepackage{float}

\usepackage{adjustbox}
\usepackage{booktabs}
\usepackage{bm}
\usepackage{subcaption}

\usepackage{amsmath,amssymb,amsthm,mathtools}
\usepackage{breqn}
\usepackage{bm}
\usepackage[capitalize]{cleveref} % must go after other package loads

\newcommand{\suchthat}{\;\ifnum\currentgrouptype=16 \middle\fi|\;}

\newcommand{\vizhints}{\textsc{VisualHints}}

\usepackage{adjustbox}
\usepackage{booktabs}
\usepackage{bm}

\usepackage[switch]{lineno} % Adding line nos. on recommendation of AAAI email

\newlength\myindent
\begin{document}
	% \title{Take the Hint: A Multimodal Approach towards Contextualized Visual Attention in Text-based games}
	\title{\vizhints: A Visual-Lingual Environment for\\ Multimodal Reinforcement Learning}
	
	\author{
		Thomas Carta\textsuperscript{\rm 1}\footnote{Both authors contributed equally},
		Subhajit Chaudhury\textsuperscript{\rm 2}$^*$,
		Kartik Talamadupula\textsuperscript{\rm 2},
		Michiaki Tatsubori\textsuperscript{\rm 2}
	}
	\affiliations{
		% Affiliations
		\\
		\textsuperscript{\rm 1}Ecole Polytechnique
		\textsuperscript{\rm 2}IBM Research 
		
		\texttt{thomas.carta@polytechnique.edu},~ \texttt{subhajit@jp.ibm.com},~ \texttt{krtalamad@us.ibm.com},~ \texttt{mich@jp.ibm.com}
	}
	\maketitle
	
	%\linenumbers % Adding line nos. on recommendation of AAAI email
	% SUBMISSION ONLY - REMOVE FOR CAMERA READY
	
	\begin{abstract}
		
		We present \vizhints, a novel environment for multimodal reinforcement learning (RL) involving text-based interactions along with visual hints (obtained from the environment). Real-life problems often demand that agents interact with the environment using both natural language information and visual perception towards solving a goal. However, most traditional RL environments either solve pure vision-based tasks like Atari games or video-based robotic manipulation; or entirely use natural language as a mode of interaction, like Text-based games and dialog systems. In this work, we aim to bridge this gap and unify these two approaches in a single environment for multimodal RL. We introduce an extension of the TextWorld cooking environment with the addition of visual clues interspersed throughout the environment. The goal is to force an RL agent to use both text and visual features to predict natural language action commands for solving the final task of cooking a meal. We enable variations and difficulties in our environment to emulate various interactive real-world scenarios. We present a baseline multimodal agent for solving such problems using CNN-based feature extraction from visual hints and LSTMs for textual feature extraction. We believe that our proposed visual-lingual environment will facilitate novel problem settings for the RL community.
		
	\end{abstract}
	
			\begin{figure*}[htb]
		
		\centering
		\includegraphics[width=0.8\textwidth]{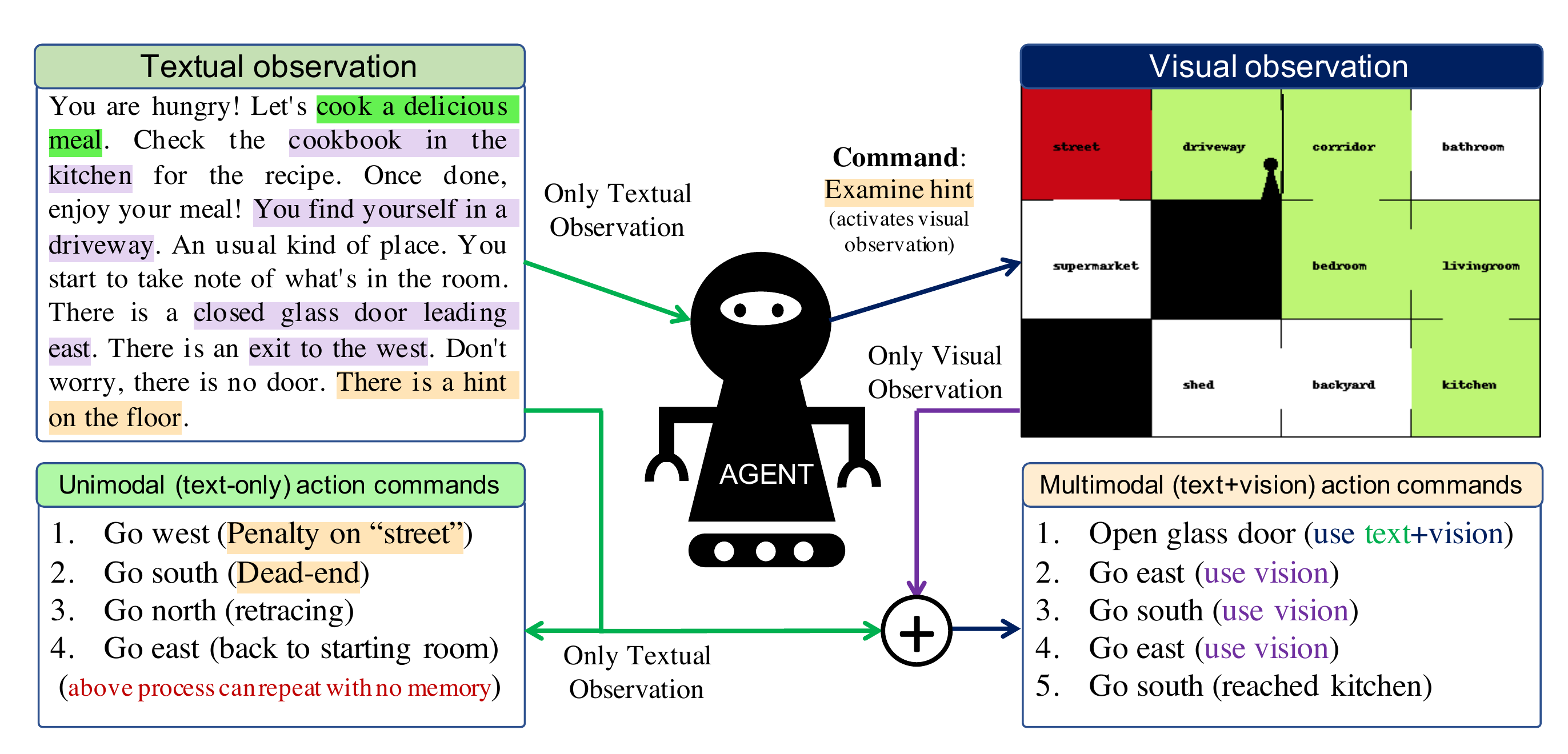}
%		\caption{An overview of the \vizhints\ domain. The policy uses both image and text from the environment to produce the next action. Step 1 shows no visual hint available (black image input to the CNN). The textual features from the GRU unit are used to generate the ``examine hint'' command; then, in Step 2, the visual hint is revealed to the visual processing module. Using the text alone, the agent can venture into two different directions: north, or east. With the visual hint, however, the ``go north'' command is generated by the agent.}
		
		\caption{An overview of our \vizhints\ environment. The agent starts at the ``driveway'' where it only receives the textual observation. From this textual observation, the agent can issue the command ``examine hint'' to obtain the visual hint shown on the top-right. If the agent does not use the visual hint and simply tries to use textual observation, then it is likely to venture in a wrong path ``driveway''$\rightarrow$``street''$\rightarrow$``supermarket'' due to partial observability. Using both text and visual observation (after picking up the hint), the agent has a better context to reach the ``kitchen''. However, our environment design necessitates both visual and textual information, which requires solving novel problem settings compared to existing works for multimodal RL. }
		\label{fig:intro_fig}
		
	\end{figure*}
	
	\section{Introduction}
	\label{sec:introduction}
	
	Reinforcement Learning~(RL) methods~\cite{sutton1998reinforcement} learn action policies for agents to achieve a given task by maximizing a guiding reward signal. Recently, there has been great interest in the RL community in using deep learning (DL) based methods, following up on the success of deep RL~\cite{Mnih2015, schulman2015trust} in vision-based high dimensional environments like Atari~\cite{bellemare13arcade, 1606.01540} and Mujoco~\cite{todorov2012mujoco}. Similarly, RL methods are increasingly being used to solve real-world problems involving sequential decision making from natural language descriptions. Typical applications in this domain include  text-based game agents~\cite{narasimhan2015language, he2016deep, yuan2018counting}, chatbots~\cite{serban2017deep}, and personal conversation assistants~\cite{dhingra2017towards}. While these methods focus either on purely image-based or purely text-based observations respectively, there is a lack of methods that address joint multimodal learning from both these domains. In this paper, we present a visual-lingual environment to facilitate action policy learning via multimodal RL.

	Humans interact with their environments using various modalities of input. For example, let us say that a person wants to drive up to a certain location. For that purpose, they use visual inputs for determining the direction of navigation from a map. However, an announcement on the car radio warns of traffic congestion on the designated route. In this situation, humans can replan and find the new optimal path (with respect to time) using multiple input modalities. However, previous environments for RL primarily provide observations from a single input modality -- either image or text. For such RL methods to generalize and scale to real-world applications, it is necessary to handle multimodal inputs for solving the required task. The target environment should be such that both the textual and visual modalities are required to solve the problem; unimodal input is not enough to solve the problem successfully.
	
	% \smallbreak
	Our goal is to develop an environment that provides both image and text information for agents to interact with. There are many real-world applications for such environments. One of them is chatbots in online shopping websites that can use pictures and textual descriptions from the user to enhance their ability to advise the customer and to carry on a conversation in a natural way. Another example is in construction areas, where agents have to move equipment or materials following the verbal instructions of the foreman, and a map of the construction site. These examples show the necessity of building agents that can extract and relate textual and visual clues; prioritize them; and use them at the right time.
	
	% \smallbreak
	
	In this work, we present \vizhints, a visual-lingual environment for multimodal, natural interactions. Such interactions should help agents build representations of the environment's state faster and more accurately than a unimodal approach. In this work, we have expanded the natural language problem to a problem that couples both natural language and vision. We used \textit{Cooking Game}, a text-based game (TBG) generated by TextWorld~\cite{cote2018textworld} (a sandbox developed by Microsoft) as our base environment; we added visual clues to this base. A visual clue is a helpful hint that agents can very likely expect from humans in real-world interaction contexts. We decided to impose two constraints on the visual clues: (i) preservation of the partial observability of the environment (to maintain the nature of the game as a Partially Observable Markov Decision Process (POMDP)); and (ii) different levels of granularity with respect to the visual indicators (to reveal smaller or larger parts of the environment).
	
	% Indeed, a visual clue seems to be the type of helpful information that an agent could obtain from a human in a real interaction context. We decided that the visual clue should respect two constraints: first no loss of partial observability of the environment (to maintain the partially observable Markov decision process nature of the game), second delivering different levels of visual indications( to reveal a more or less large part of the environment). 
	% \smallbreak
	
	To meet the above constraints for the \vizhints\ environment, we generated maps for all the preexisting \textit{Cooking Games} in the form of a floor plan; with textual indications (clues) that allow each map to be interpreted, and for which the difficulty level can be set. The map-clue (visual-lingual) pair allows us to encode varying degrees of abstract information, ranging from a factual and spatial description of the environment to a reward oriented description of the game. The agent thus has to learn the interactive meaning between vision and text to obtain reliable information. No previous work on TBGs has considered such a setting: we therefore developed a program that encapsulates TextWorld's games, and generates clues and maps related to each game. We have also built a baseline agent that can use a multimodal approach to solve the game. This agent combines a textual module inspired by the work of \citet{adolphs2019ledeepchef} and a visual module that we have designed ourselves to analyze the floor plan.

	\smallbreak
	
	\noindent The main contributions of this paper are:
	
	\begin{itemize}
		
		\item We introduce \vizhints, a visual-lingual environment for multimodal reinforcement learning that requires inputs from both textual and visual components of the input observation. Our proposed visual hints are generalizable and can be generated automatically across domains. In this paper, we show  the Cooking Games task from TextWorld with different structures of the visual hints.
		
		\item We propose a baseline agent for solving the Cooking World task of preparing and eating a meal that can use both textual input using LSTM features, and visual indications using CNNs to successfully solve a problem.
		
		\item We analyse how the visual model interacts with the input image to understand a map and its topology based on a pre-training task.
		
	\end{itemize}

	\noindent The paper is organized as follows. We first present the design of \vizhints, a multimodal learning environment using the image and textual observations in the context of a Partially Observable Markov Decision Process (POMDP). We present the architecture of a baseline multimodal agent with an Advantage Actor Critic (A2C) based model-free algorithm to solve such a problem. We then explain our methodology to build the map and the puzzle; to train the visual part of the model; and to train the full model. Following this, we analyze the results of pretraining the visual module of the agent. Finally, we evaluate the baseline multimodal agent in comparison with previous state-of-the-art text-only methods.
	
	%Similarly, a well-defined reward signal might not be always available for policy learning in artificial agents as well. In such cases, 
	
	%\begin{figure}[tb]
	%	\centering
	%	\includegraphics[width=8cm, height=3.6cm]{figures/Idea.pdf}
	%	\caption{.}
	%	\label{fig:idea}
	%\end{figure}

	\section{Related Work} %% Please check here!!
	
	Given the nature of our contribution, we examine related works along the two modalities that we consider: text-based games and image-based interactive environments.
	
	\subsection{Text-Based Games}
	
	LSTM-DQN~\cite{narasimhan2015language} was the first work to tackle text-based games using RL; it uses deep Q-learning-based reinforcement learning to jointly learn state representations and action policies using game rewards as feedback. The Textworld~\cite{cote2018textworld} framework enabled the generation of many different configurations of text-based games. Using this framework, text descriptions can be mapped into vector representations that capture the semantics of game states. 
	LSTM-DQN assumed some structure in the output command setting; this was tackled by DRRN~(Deep Reinforcement Relevance Network) by addressing a natural language action space for generating action commands~\cite{he2016deep}. This method uses separate embedding vectors for action and state spaces: an interaction function combines them for input to the Q-function module.  
	
	LSTM-DRQN~\cite{yuan2018counting} is one of the state-of-the-art methods for the TextWorld Coin Collector domain: it was proposed in order to address the issue of partial observability. This method processes textual observations for the recurrent policy to generate a vector representation that estimates Q-values for all verbs $Q(s, v)$ and objects $Q(s, o)$. This work also proposes discovery bonuses for generalizing better on unseen games. There were two types of bonuses: a cumulative counting bonus which gradually converges to 0; and an episodic discovery bonus which encourages only first-time discovery of unseen states. 
	
	There are some prior efforts~\cite{fulda2017can,zahavy2018learn} that deal with pruning action tokens which do not agree with the current state's context; this brings about accelerated convergence of the RL agent. \citet{fulda2017can} proposed a method for affordance extraction via word embeddings trained on a Wikipedia corpus: they showed that previously intractable search spaces can be efficiently navigated when word embeddings are used to identify context-dependent affordances. AE-DQN~(Action-Elimination DQN)---which is a combination of a Deep RL algorithm with an action eliminating network for sub-optimal actions---was proposed by \citet{zahavy2018learn}. The action eliminating network is trained to predict invalid actions, supervised by an external elimination signal provided by the environment. More recent methods~\cite{adolphs2019ledeepchef, ammanabrolu2018playing, ammanabrolu2020graph,yin2019learn, adhikari2020learning} use different heuristics to learn better state representations for efficiently solving complex TBGs.
	
	\subsection{Image-Based Interactive Environments}
	\label{subsec:related_image_environments}
	
	% \textcolor{red}{<write here about AI2THor and related environments>} . 
	In the past couple of years, there has also been a proliferation of environments that focus on the visual modality, and represent the environment in terms of an image or video feed. Examples of such environments include AI2-THOR~\cite{kolve2017ai2}, Habitat~\cite{savva2019habitat}, ALFRED~\cite{ALFRED20}, and AllenAct~\cite{AllenAct}. These environments typically feature an agent that is situated in a home-like environment, and must accomplish tasks by using its perception capabilities. However, these environments have a fixed set of action commands to choose from based on the visual observation and scenario. For example, near the object ``TV'', the  interactive commands are limited and structured, e.g. \textit{turn on} or \textit{turn off}. Additionally, some of these environments offer the agent the ability to {\em sub-sample} the space in a specific modality: for e.g., agents in AI2-THOR can choose to receive state information purely as text-based logical state descriptors, thus greatly cutting down on the complexity of the inherent task. In contrast, our environment provides natural language action commands that are much more difficult to handle due to the large action space. Moreover, we make sure that for games in our environment, both visual and textual modalities are required in order to arrive at a solution; this is not guaranteed by the other environments.

	\section{Problem Setting: Multimodal RL}
	\label{sec:mlrl}
	
	Text-based games (TBGs) are sequential decision-making problems that we aim to solve using model-free RL algorithms. These environments form POMDPs (Partially Observable Markov Decision Processes) because the agent receives partial information about its surroundings from the environment. In this paper, we aim to add visual information in the form of a map of the environment to help the agent in its quest to solve the game. In a typical Cooking World TBG with a map, the POMDP is defined by $(S, A, T, \Omega, O, R, \gamma$); where $S$ is a non-empty finite set of possible states of the system to be controlled,  
	% which has global knowledge of the environment
	$A$ is the Action space, i.e., the non-empty finite set of actions that can be performed to control the system. In our case, we have natural language action commands which are a combination of Verbs $\times$ Adjectives $\times$ Nouns. $T: S\times S\times A \rightarrow [0;1]$ is the transition function of the system (which we do not know in a model-free setting); and $\Omega$ is a set of observable symbols.  $O$ denotes the observation, which is a part of the state that is observed by the agent. Typically in TBGs, only a textual description $O_t^{text}$ of the world forms the observation; whereas in visual environments like Atari, only image-based observations $O_t^{image}$ are obtained. In \vizhints, we necessitate that both visual and textual information be used for action command generation. Thus, we define a policy $\pi:O_t^{text}, O_t^{image}\rightarrow A_t$ that generates trajectories in this environment. The goal of multimodal RL is to find a policy $\pi$ by maximizing the expected reward $J(\pi) = \mathbb{E}_{\tau\sim\pi}[ \sum_{t\in\tau} \gamma^{t}r_{t} ]$.

	\begin{figure}[tb]
		\centering
		\includegraphics[width=0.45\textwidth]{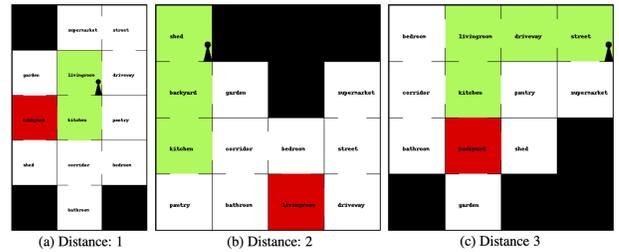}
		\caption{Example of visual hints with various distance of puzzle values.}
		\label{fig:cooking_game_map}
	\end{figure}

	% \section{Our Method}
	
	\section{The \vizhints\ Environment}
	\label{sec:environment}
	
	As shown in Figure~\ref{fig:intro_fig}, our \vizhints\ environment provides a visual clue along with a short hint text, which can be used for the multimodal RL setting described in Section~\ref{sec:mlrl}.  Such multimodal observations can give information to the agent to efficiently explore and solve the game. We add various modes in our \vizhints\ environment that we describe in detail in the following section. 
	
	While performing correct action command prediction is paramount for solving RL tasks, it is also important in real-world applications to learn what action commands should not be generated to prevent fatally dangerous situations. To simulate this effect, we have added the notion of ``death rooms'', which are rooms that lead to the death of the agent, with large negative rewards. These rooms are similar in spirit to the notion of ``dead end'' states that make a problem ``probabilistically interesting''~\cite{little2007probabilistic}. Such death rooms can also be placed on the direct path between the source and destination rooms in the environment, in which case the agent has to find an alternative route that may not be the shortest path, but is safe.
	
	A game begins with some information on the board, which the agent has to read by issuing the ``read board'' command. The board indicates the presence and location of the death room. The board is always placed in the first room to ensure that the player always has a chance to know where the death room is before encountering it. Somewhere during the course of the game, the agent finds a \textit{hint} which is made up of textual and visual information (floor layout) as shown in Figure~\ref{fig:intro_fig}. The text gives some information on how to interpret the map. For instance, the agent can find this map alongside a textual clue: 'take the ingredients in the kitchen, the supermarket, and cook in the kitchen, and avoid the death room which is the bathroom'.  In the example of Figure~\ref{fig:cooking_game_map}, the player is in the living room. The death room is in red, and the pathway leading to the cooking location is in green.
	
	%\begin{figure}[tb]
	%    \centering
	%    \includegraphics[width=0.3\textwidth]{figs/cooking_game_map.png}
	%    \caption{An example of visual hint that can be found in the game containing death room as ``bathroom''.}
	%    \label{fig:cooking_game_map}
	%\end{figure}
	
	\smallbreak

	\subsection{Generating Hints from TextWorld}
	\label{subsec:generating_hints_from_tw}
	
	We use the cooking games used by \citet{adolphs2019ledeepchef} as the base games upon which we add the visual hints for multimodal RL. In order to implement the automatic and general addition of visual hints and textual clues to the existing textual observation space, we create an extra layer between the agent and the TextWorld game. The visual hints are generated alongside textual clues. A total of $7$ different modes can be tuned to automatically generate the map, which we describe here.

	\subsubsection{Distance of Puzzle} 
	
	This mode controls the distance between the room where the hint is found and the primary destination (the cooking location). This emulates the difficulty of learning the visual component of the RL model. A high value would correspond to high difficulty in learning the navigation commands, whereas a low value would correspond to a simpler learning setting. Figure~\ref{fig:cooking_game_map} shows variations on this factor.
	
	\subsubsection{Death Room} 
	
	Another decision factor in the environment is the presence of the death room. It is possible that a death room may not be added to a specific game instance: for example, if the game has only one room, or if all rooms are necessary to arrive at a solution. Adding a death room emulates the factor of safety in real-world tasks,  and makes the final task more difficult to achieve.
	
	%\textbf{max_number_inaccessible_rooms} number of rooms that are inaccessible because of the death room
	
	% \texttt{room_name} if you want to put a name on rooms
	
	\subsubsection{Color Path}
	
	This allows for the coloring of the path between the agent's current location (the location where the hint is found) and the primary goal (the cooking location). This helps the RL agent by providing a color-based visual guide that would be otherwise difficult to obtain.
	
	% Whether to color the way between the current room of the agent (where the hint is found) and the primary goal (the cooking place). This helps the RL agent by providing a color-based guide that would be otherwise difficult to establish as guiding visual signals. 
	
	\begin{figure}[tb]
		\centering
		\includegraphics[width=0.45\textwidth]{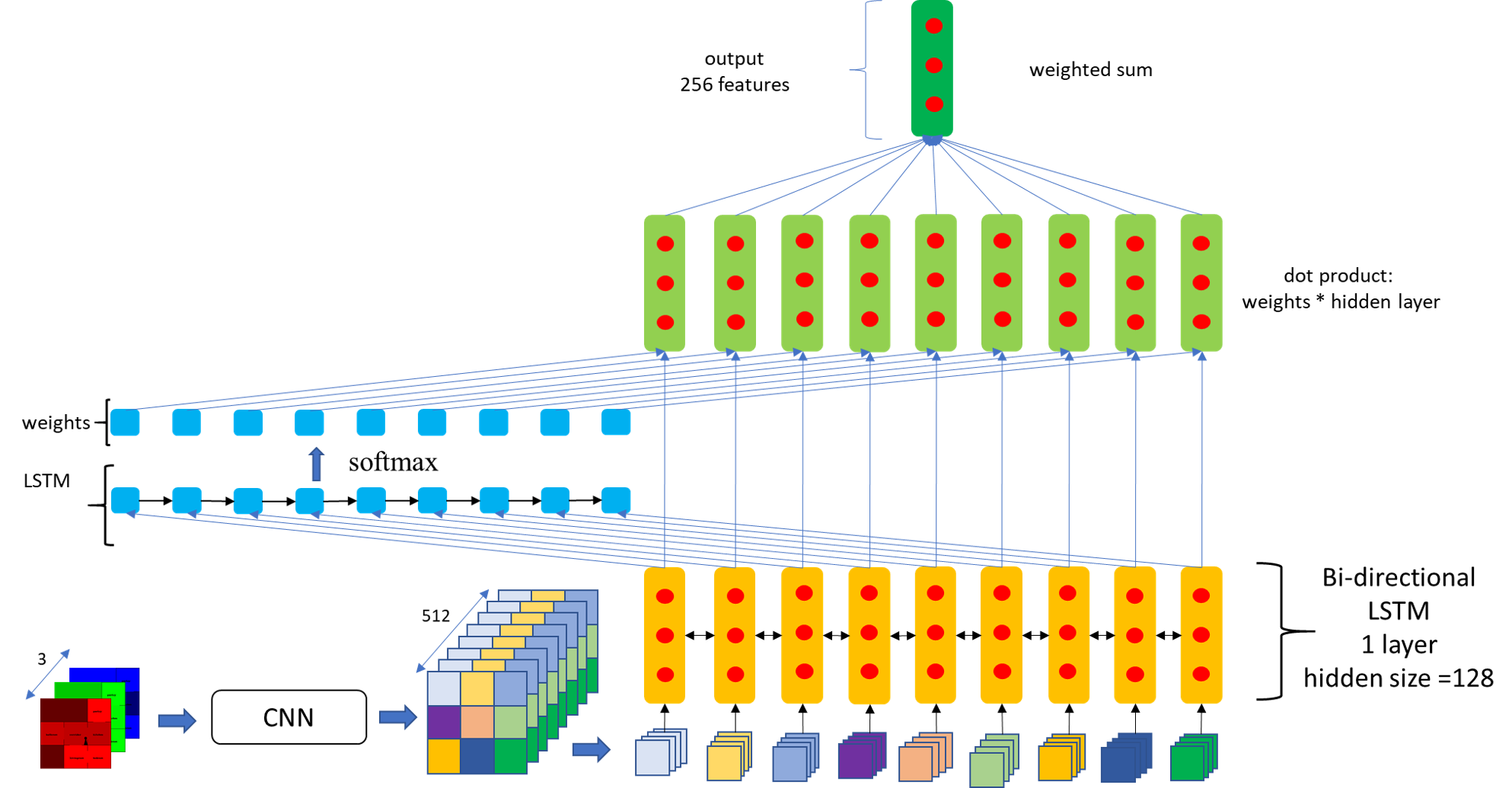}
		\caption{Architecture of the CNN-LSTM based visual observation parsing module of the agent.}
		\label{fig:cnn_lstm_architecture}
	\end{figure}
	
	\subsubsection{Name Type}
	
	These following options enable the control of the visualization of each individual room for the sake of giving control to the learning type.
	
	\begin{itemize}
		\item \texttt{literal}: the established name of the room (‘pantry’, ‘kitchen’)
		\item \texttt{random_numbers}: the name of a room is a random number (all rooms have different numbers)
		\item \texttt{room_importance}: the name of a room is a number denoting the importance of that room in that specific game 
		\begin{itemize}
			\item 1 if it is the primary goal (the cooking location)
			\item 2 if it is a secondary goal (contains an ingredient)
			\item 0 otherwise
		\end{itemize}
	\end{itemize}
	
	\subsubsection{Draw Passages}
	
	This option determines whether passages (open or closed) are drawn between adjacent rooms on the map. If this information is provided, the agent can obtain all of the required navigational information purely from the visual hint; otherwise, the agent will need to read the corresponding text as well in order to find out which direction is accessible.
	
	% Whether to show the passages between the adjacent rooms on the map. If this information is provided, then the agent can obtain entire navigation information from the visual hint, otherwise, it needs to read the corresponding text as well to find out which direction is accessible.
	
	% \smallbreak

	\subsubsection{Draw Player}
	
	This option determines whether the agent's location on the map is included as part of the visual hint.
	
	% Whether to show the location of the player on the map.
	%\texttt{random_place} if you want to obtain a random place for the hint respecting \texttt{distance_of_puzzle}, if \textit{False} the hint will always be placed on the same room respecting \texttt{distance_of_puzzle}
	%\texttt{mask} put a mask on the picture to only have the colorway visible
	
	\subsubsection{Clue First Room}
	
	This setting forces the clue to be placed in the first room, irrespective of the value of ``distance of puzzle''. This reduces the effort the agent has to spend in searching for the hint during the initial stages without any visual signal.
	
	% force the clue to be in the first room irrespective of the ``$distance_of_puzzle$'' value. This reduces the effort of the agent to search for the hint at the first stages without any visual signal.  
	
	\smallbreak
	
	\noindent Along with the visual hint, two text-based clues are also provided: one at the beginning that explains where the death room is located; and the second a text-based hint that accompanies the visual hint. We also add difficulty to these textual clues (described in detail in the supplementary material).
	
	Based on the above description, a large variety of problem settings can be generated by our \vizhints\ environment for multimodal RL. These tasks serve as a proxy for real-world RL tasks that utilize text and visual inputs in the delayed reward setting. To the best of our knowledge, there are no previous multimodal RL environments that provide such a detailed level of granularity and control. Furthermore, our environment is readily portable to other domains within TextWorld as well. We therefore hope that our \vizhints\ environment will serve as a benchmark for the multimodal RL community.

	\begin{figure}[tb]
		\centering
		\includegraphics[width=0.45\textwidth]{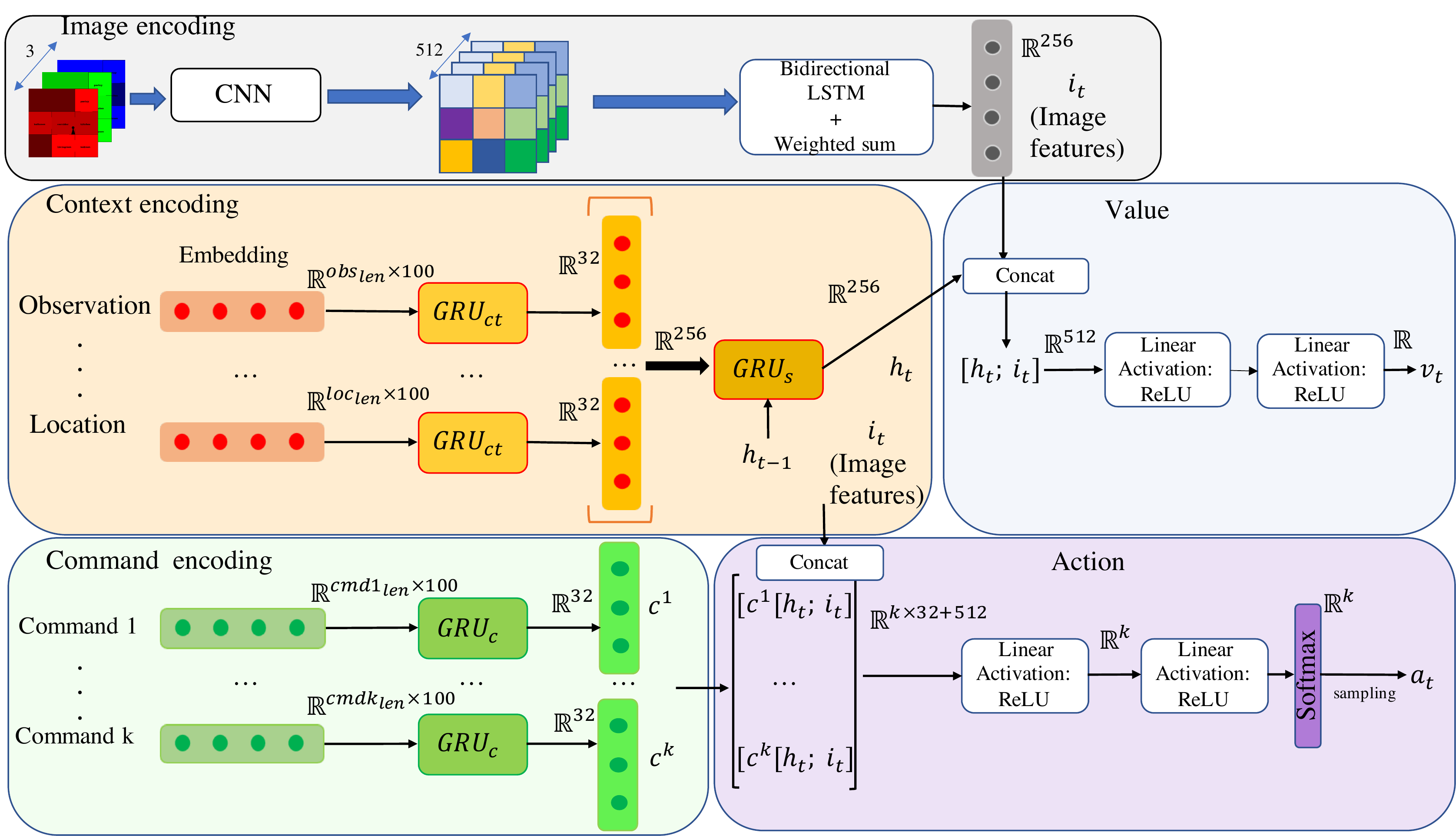}
		\caption{The final multimodal RL policy architecture that utilizes both textual and visual information.}
		\label{fig:overall}
	\end{figure}
	
		%%%%%%%%%%%%%%%%%%%%%%%%%%%%%%%%%%%%%%%%%%

	%	\newpage
	
	\begin{figure*}[!htb]
		\centering
		\begin{subfigure}[b]{0.24\textwidth}
			\includegraphics[trim={0mm 0mm 0mm 0mm},clip,width=0.99\linewidth]{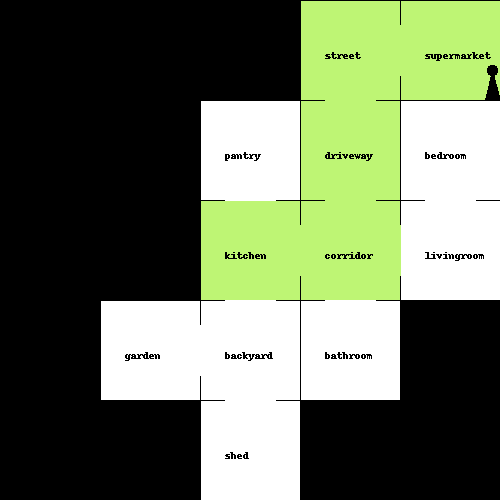}
			\caption{Game 27}
		\end{subfigure}%
		\begin{subfigure}[b]{0.24\textwidth}
			\includegraphics[trim={0mm 0mm 0mm 0mm},clip,width=0.99\linewidth]{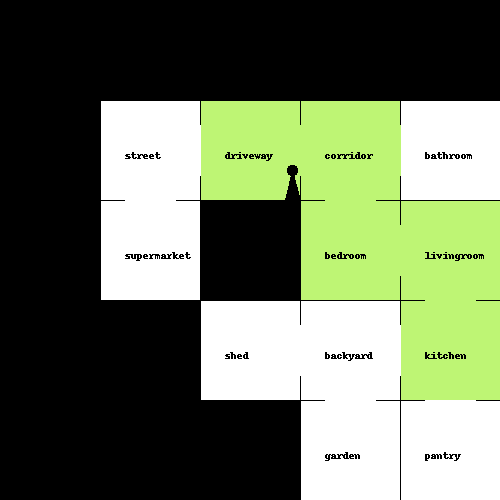}
			\caption{Game 35}
		\end{subfigure}%
		\begin{subfigure}[b]{0.24\textwidth}
			\includegraphics[trim={0mm 0mm 0mm 0mm},clip, width=0.99\linewidth]{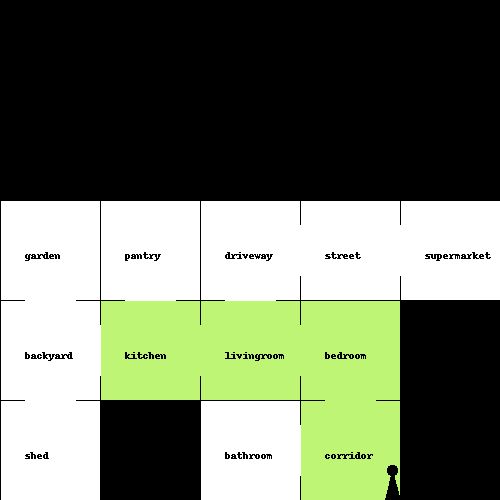}
			\caption{Game 40}
		\end{subfigure}%
		\begin{subfigure}[b]{0.24\textwidth}
			\includegraphics[trim={0mm 0mm 0mm 0mm},clip, width=0.99\linewidth]{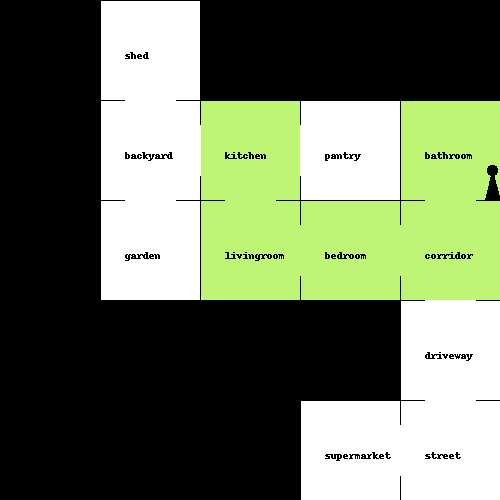}
			\caption{Game 45}
		\end{subfigure}%
		\caption{Showing different configurations of the VisualHints used in the cooking game. The goal of the agent is to reach the kitchen room, where the agent has to prepare the meal and eat the meal to win the game.}\label{fig:simple}
	\end{figure*}
	
	% Death room
	\begin{figure*}[!htb]
		\centering
		\begin{subfigure}[b]{0.24\textwidth}
			\includegraphics[trim={0mm 0mm 0mm 0mm},clip,width=0.99\linewidth]{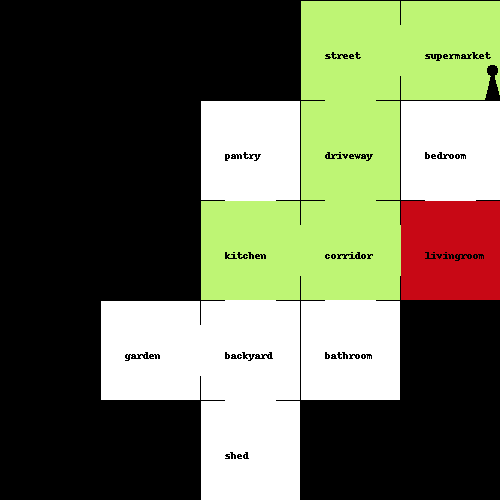}
			\caption{Game 27}
		\end{subfigure}%
		\begin{subfigure}[b]{0.24\textwidth}
			\includegraphics[trim={0mm 0mm 0mm 0mm},clip,width=0.99\linewidth]{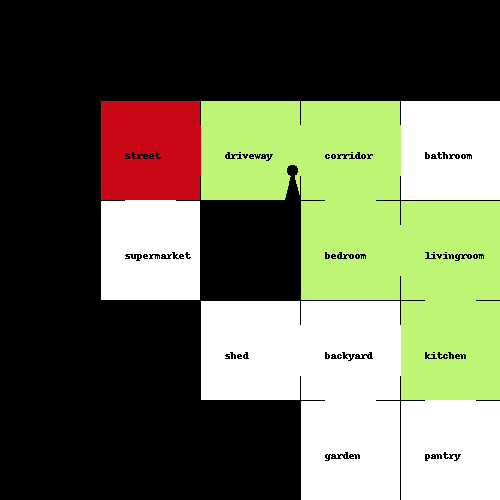}
			\caption{Game 35}
		\end{subfigure}%
		\begin{subfigure}[b]{0.24\textwidth}
			\includegraphics[trim={0mm 0mm 0mm 0mm},clip, width=0.99\linewidth]{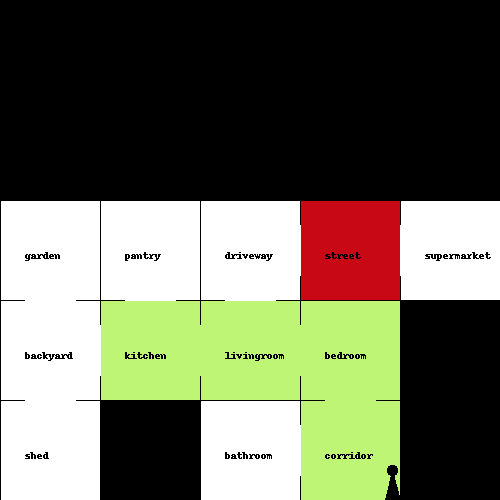}
			\caption{Game 40}
		\end{subfigure}%
		\begin{subfigure}[b]{0.24\textwidth}
			\includegraphics[trim={0mm 0mm 0mm 0mm},clip, width=0.99\linewidth]{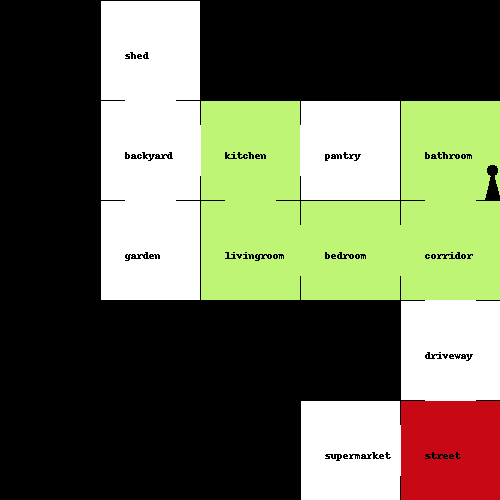}
			\caption{Game 45}
		\end{subfigure}%
		\caption{Adding death room to VisualHints to emulate the role of safety in the real world. The modified goal of the agent is to reach the kitchen room while avoiding going to the death-room. The agent dies and the game is over if the agent visits the death-room.}\label{fig:death}
	\end{figure*}
	
	% Partial room
	\begin{figure*}[!htb]
		\centering
		\begin{subfigure}[b]{0.24\textwidth}
			\includegraphics[trim={0mm 0mm 0mm 0mm},clip,width=0.99\linewidth]{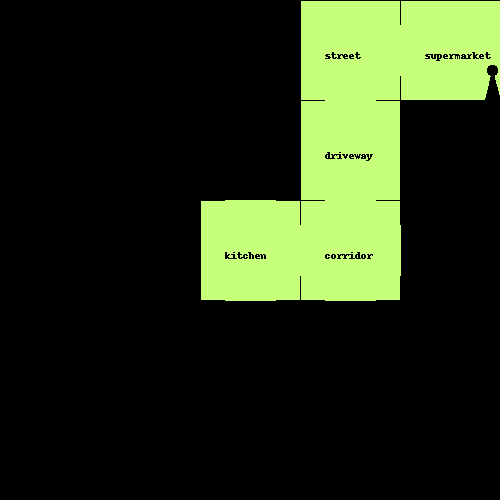}
			\caption{Game 27}
		\end{subfigure}%
		\begin{subfigure}[b]{0.24\textwidth}
			\includegraphics[trim={0mm 0mm 0mm 0mm},clip,width=0.99\linewidth]{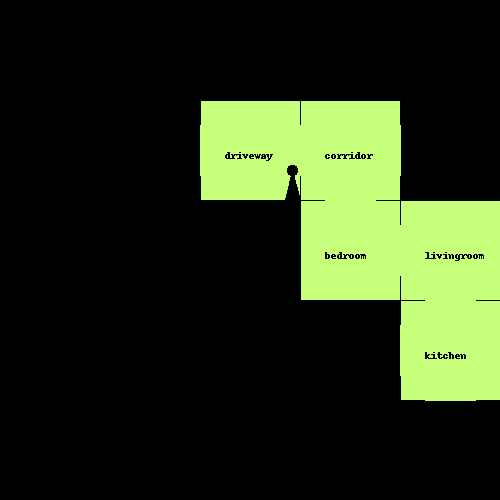}
			\caption{Game 35}
		\end{subfigure}%
		\begin{subfigure}[b]{0.24\textwidth}
			\includegraphics[trim={0mm 0mm 0mm 0mm},clip, width=0.99\linewidth]{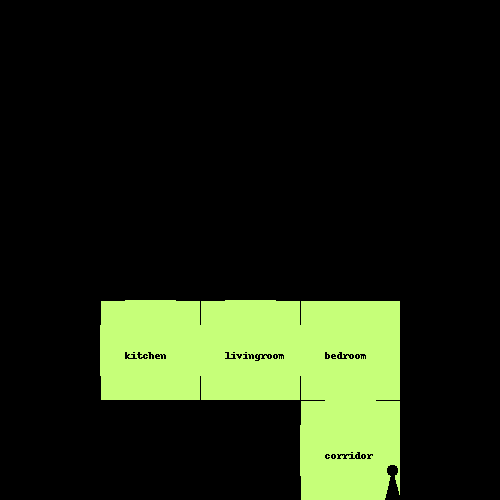}
			\caption{Game 40}
		\end{subfigure}%
		\begin{subfigure}[b]{0.24\textwidth}
			\includegraphics[trim={0mm 0mm 0mm 0mm},clip, width=0.99\linewidth]{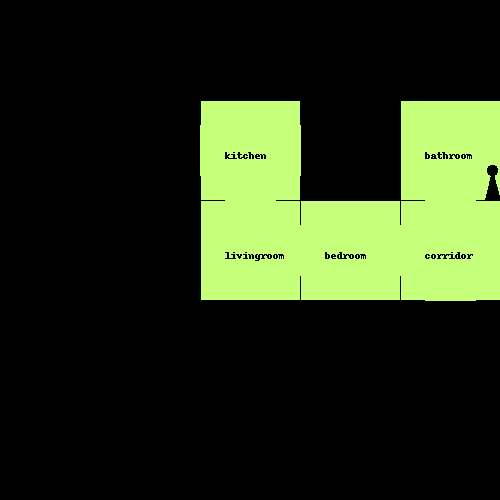}
			\caption{Game 45}
		\end{subfigure}%
		\caption{Showing masked configuration of the game where only the pathway to the kitchen is highlighted and other details are masked. This will make it easier for the agent to reach the target which is the kitchen in this case. However, even the death room is masked and hence it is dangerous for the agent to deviate from the shown green pathway.}\label{fig:mask}
	\end{figure*}

	% room importance
	\begin{figure*}[!htb]
		\centering
		\begin{subfigure}[b]{0.24\textwidth}
			\includegraphics[trim={0mm 0mm 0mm 0mm},clip,width=0.99\linewidth]{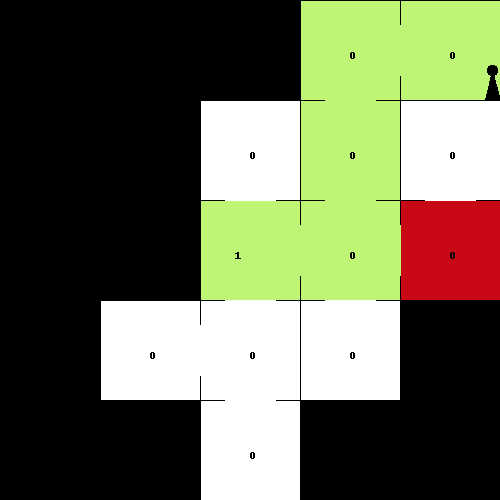}
			\caption{Game 27}
		\end{subfigure}%
		\begin{subfigure}[b]{0.24\textwidth}
			\includegraphics[trim={0mm 0mm 0mm 0mm},clip,width=0.99\linewidth]{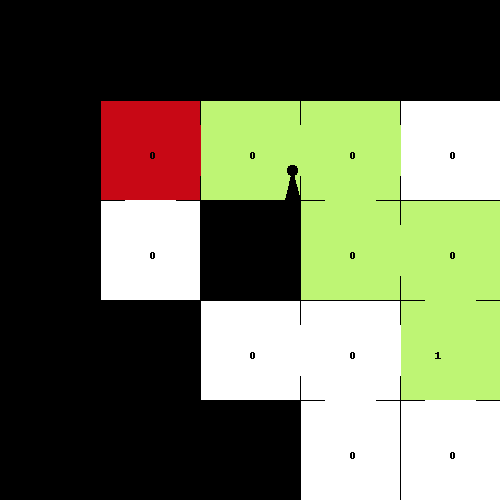}
			\caption{Game 35}
		\end{subfigure}%
		\begin{subfigure}[b]{0.24\textwidth}
			\includegraphics[trim={0mm 0mm 0mm 0mm},clip, width=0.99\linewidth]{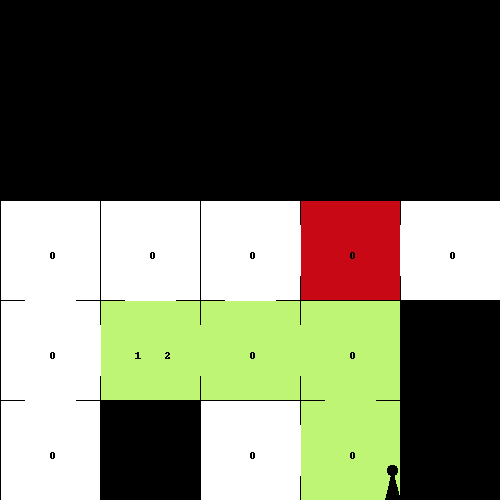}
			\caption{Game 40}
		\end{subfigure}%
		\begin{subfigure}[b]{0.24\textwidth}
			\includegraphics[trim={0mm 0mm 0mm 0mm},clip, width=0.99\linewidth]{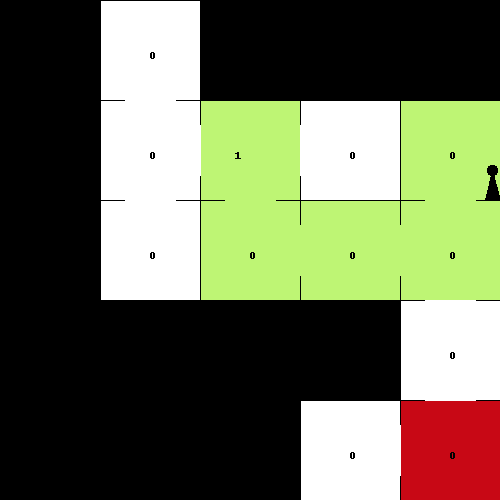}
			\caption{Game 45}
		\end{subfigure}%
		\caption{In this mode, the name of the room is masked and only the room importance is displayed. For the primary target~(kitchen) the room importance is kept at 1 and for secondary targets (like rooms where the ingredients need to be picked from) are shown with importance of 2. All other rooms are shown with importance of 0.}\label{fig:importance}
	\end{figure*}
	
	\begin{figure*}[!htb]
		\centering
		\begin{subfigure}[b]{0.24\textwidth}
			\includegraphics[trim={0mm 0mm 0mm 0mm},clip,width=0.99\linewidth]{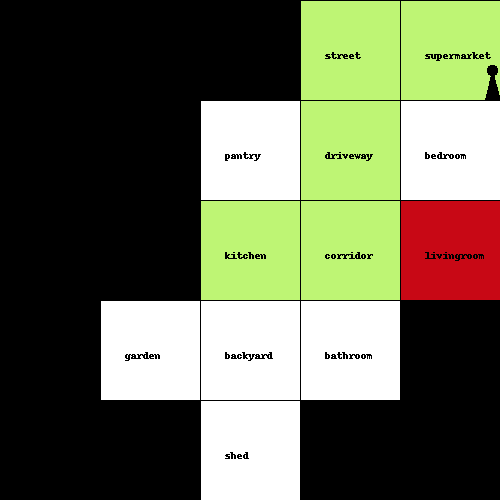}
			\caption{Game 27}
		\end{subfigure}%
		\begin{subfigure}[b]{0.24\textwidth}
			\includegraphics[trim={0mm 0mm 0mm 0mm},clip,width=0.99\linewidth]{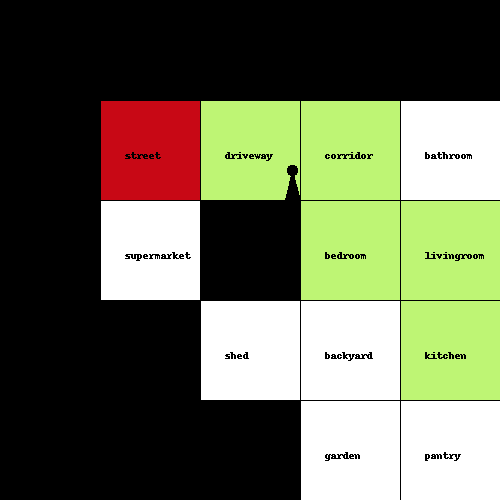}
			\caption{Game 35}
		\end{subfigure}%
		\begin{subfigure}[b]{0.24\textwidth}
			\includegraphics[trim={0mm 0mm 0mm 0mm},clip, width=0.99\linewidth]{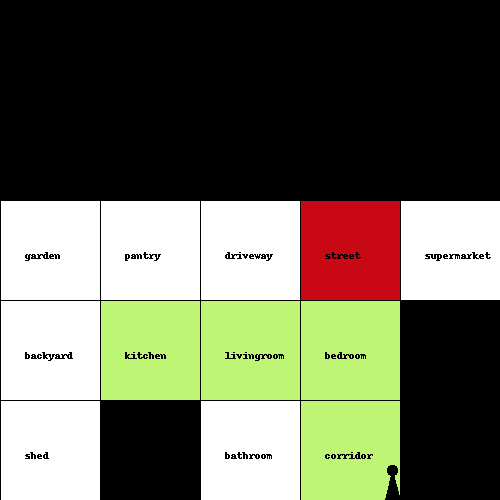}
			\caption{Game 40}
		\end{subfigure}%
		\begin{subfigure}[b]{0.24\textwidth}
			\includegraphics[trim={0mm 0mm 0mm 0mm},clip, width=0.99\linewidth]{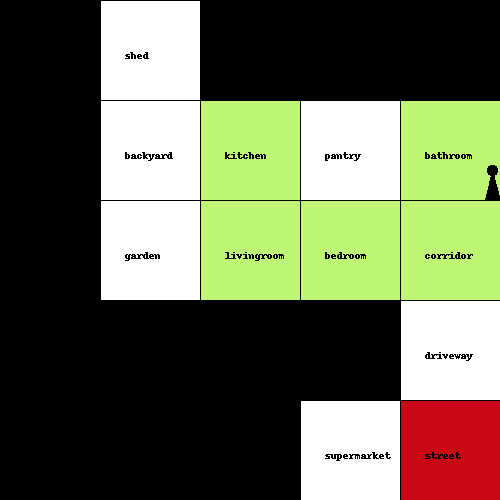}
			\caption{Game 45}
		\end{subfigure}%
		\caption{In this mode, the pathways between each room are masked and the agent cannot have a notion of which direction is available for travel only from the visual hint. In such a case, the multi-modal agent has to rely on the textual observation to ascertain which direction has an open door and thus decide to take action accordingly.}\label{fig:noway}
	\end{figure*}

	\begin{figure*}[!htb]
		\centering
		\begin{subfigure}[b]{0.24\textwidth}
			\includegraphics[trim={0mm 0mm 0mm 0mm},clip,width=0.99\linewidth]{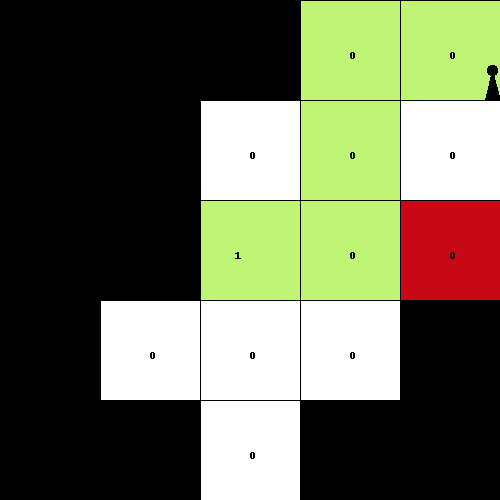}
			\caption{Game 27}
		\end{subfigure}%
		\begin{subfigure}[b]{0.24\textwidth}
			\includegraphics[trim={0mm 0mm 0mm 0mm},clip,width=0.99\linewidth]{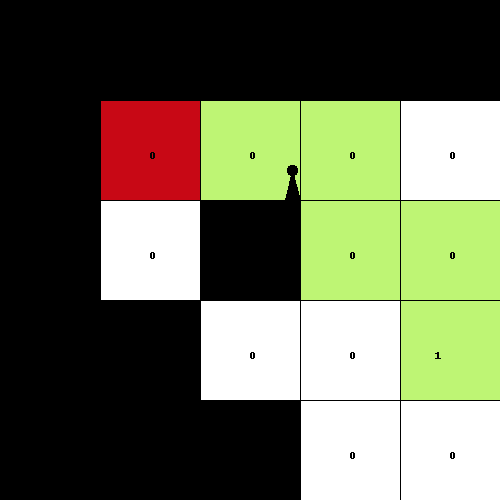}
			\caption{Game 35}
		\end{subfigure}%
		\begin{subfigure}[b]{0.24\textwidth}
			\includegraphics[trim={0mm 0mm 0mm 0mm},clip, width=0.99\linewidth]{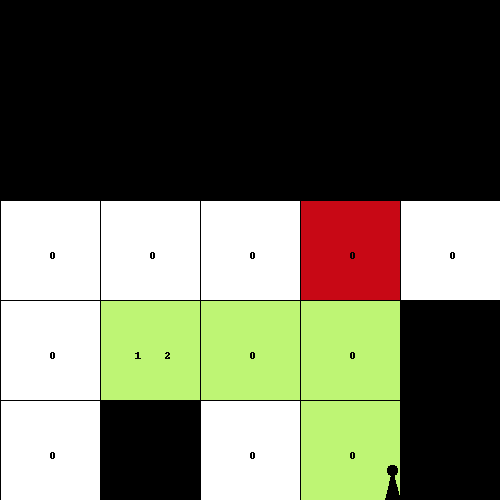}
			\caption{Game 40}
		\end{subfigure}%
		\begin{subfigure}[b]{0.24\textwidth}
			\includegraphics[trim={0mm 0mm 0mm 0mm},clip, width=0.99\linewidth]{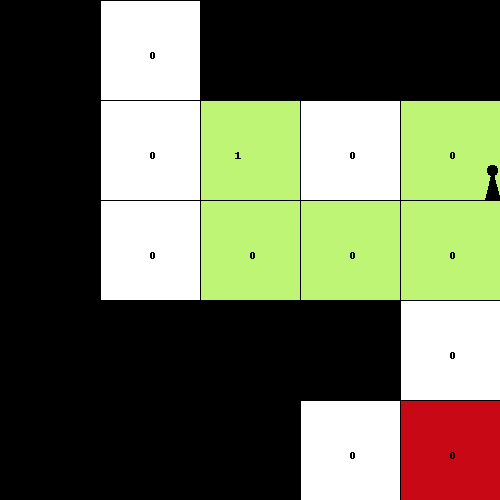}
			\caption{Game 45}
		\end{subfigure}%
		\caption{This mode combines the previous two modes where the pathways between each room are masked in addition to the roomnames being replaced with room importance.}\label{fig:noway_importance}
	\end{figure*}

	\subsection{Various operating modes of \vizhints}
\label{sec:vizhints}

In addition to the various game modes that we discussed above, we additionally show visual hints obtained under various settings of our \vizhints\ environment, in order to demonstrate the variety of situations that can be emulated by our proposed multimodal environment. Figure~\ref{fig:simple} shows a typical visual hint that can help the agent to reach the final goal viz. the kitchen. Figure~\ref{fig:death} introduces the concept of death rooms, where the agent has to be careful not to venture into dangerous areas to prevent itself from dying and ending the game without achieving the final goal. Figure~\ref{fig:mask} masks all irrelevant information about rooms to which travel is not necessary in order to solve the games, making the task of visual learning easier. Figure~\ref{fig:importance} masks the true names of the rooms and adds room importance as a proxy for the names. Figure~\ref{fig:noway} removes the visual information of the pathway between rooms such that the agent has to rely on both visual and textual information to obtain the direction of an open room, and to generate corresponding action commands. Finally,  Figure~\ref{fig:noway_importance} combines the two modes of room importance as well as the masking of visual pathways between rooms.

	%%%%%%%%%%%%%%%%%%%%%%%%%%%%%%%%%%%%%%%%%%

	\section{Baseline Model}
	\label{sec:baseline_model}
	
	We now describe a baseline model used for multimodal RL on our \vizhints\ dataset. In this work, we coupled the architecture of a textual RL agent inspired by ~\cite{adolphs2019ledeepchef} with a CNN-LSTM. The textual RL agent's architecture is used to analyze the textual part of the observation $o^{text}_{t}$, and the CNN-LSTM is used to extract information from the visual hints (the map describing the game). Advantage Actor Critic~(A2C)~\cite{mnih2016asynchronous} was used for model-free RL.
	
	% \subsection{Using CNN-LSTM to extract useful features from the images}
	
	\subsection{Extracting Useful Features from Images}
	\label{subsec:extract_useful_features_images}
	
	To extract information from the visual hints, we used a CNN followed by a bidirectional LSTM. The CNN encodes the important features of the visual hint on $512$ channels. Since the original hint has a strong underlying structure made of blocks of $100 \times 100 $ pixels, we preserve this structure. Thus, for an input picture of dimension $100n \times 100m \times 3$ with $(n,m) \in \mathbb{N}$, the CNN output is of size $p_n \times q_m \times 512$ with $(p,q) \in \mathbb{N}$. In our experiments, $(p,q) = (2,2) $.
	
	To find the link between rooms and shared properties, we pass the representation from the CNN to a bidirectional LSTM. The bidirectional LSTM encodes a spatial context of fixed size $128$ for every feature input sequence. We compute the weighted sum of all the last hidden layers of the bidirectional LSTM, and find the weights based on a second LSTM. The weights are thus related to the spatial context of the map room distribution. 
	
	In order to extract image features, we utilize a simple CNN with very few layers. The simple nature of the map, with plain colors and little text, encourages the use of simpler models (easier and faster to train) which can still capture enough details to be relevant. The CNN only consists of a convolution layer, followed by a max-pooling layer, followed by a final convolution layer.
	
	% \smallbreak
	
	\subsection{Combining Visual and Textual Features}
	\label{subsec:combine_visual_textual_observation_features}
	
	The actor and critic outputs share the part of the network that encodes information from both visual and textual observations. Observations are processed through two channels: the textual part consisting of context and command encoding, and the visual part which is responsible for encoding image features. The textual component is inspired from the work of~\citet{adolphs2019ledeepchef}. 
	% The $GRU_{s}$ cell takes the representation of the current textual context and combines it with the previous representation $h_{t-1}$ from the GRU cell to take into account the previous states.
	
	The CNN extracts local information on the map from the input channels to $512$ channels as described previously. The LSTM layer finds a link between the extracted features, resulting in a vector $i_{t} \in \mathbb{R}^{256}$. The textual and visual representations are concatenated, and we feed the concatenated inputs to single-layered multi-layer perceptrons (MLPs) for the value function and action probability distribution. The \textit{critic} part of the network takes the encoded context $h_{t} \in \mathbb{R}^{256}$ concatenated with the encoded image $i_{t}$ as input; and passes this input through an MLP with a single hidden layer of size $256$ to compute the scalar value of the game's state. The \textit{actor} part of the network takes a matrix composed of the concatenation of the encoded context and encoded image with each of the $k$ encoded commands. This vector of size $\mathbb{R}^{k \times 32 +512}$ is passed through an MLP of size $\mathbb{R}^{256}$ to compute the score vector of size $\mathbb{R}^{k}$, on which we apply the $softmax$ function to build a categorical distribution. We sample the action $a_{t}$ (one of the $k$ commands) based on this distribution.
	
	\begin{figure}[tb]
		\centering
		\includegraphics[width=0.45\textwidth]{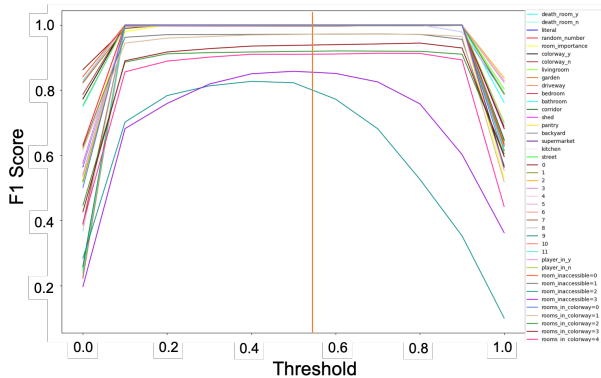}
		\caption{F1 score at various thresholds for each class in the pre-training task for the visual model.}
		\label{fig:simple_cnn_F1_score}
	\end{figure}
	
	\section{Results}
	\label{sec:results}
	
	\subsection{Experimental Details}
	Our experiments were performed on Ubuntu16.04 with TITAN X (Pascal) GPU. For the visual CNN, we used two-layers of convolutional filters with the configuration of (channels=256, kernel size=10, stride=5, padding=4) and (channels=512, kernel size=2, stride=2, padding=0) respectively. We used a Max-Pooling2D layer between the two convolutional layers.
	
	In the following, we present some results pertaining to our baseline multimodal agent in the \vizhints\ domain. We first describe the pre-training of the visual component; followed by an analysis of the CNN's output; and finally the overall performance of the baseline multimodal agent in our environment. While our results do not beat the current state-of-the-art, they are intended more as a demonstration of the potential of the \vizhints\ domain to challenge the existing best methods.
	
	\subsection{Pre-training of the CNN-LSTM}
	\label{subsec:pretraining_cnn_lstm}

	To obtain better features on the current domain, we pre-trained the visual component of the agent using $42$ questions divided into two categories:
	\begin{itemize}
		\item\textbf{Easy questions}, which do not require an understanding of the links between the different rooms on the map. For instance: Is there a `death room'? Does the name `pantry' appear?
		\item \textbf{Hard questions,} which require the relations between the different rooms. For example, ``how many rooms are blocked by the death room?''
	\end{itemize}
	Each such set of $42$ questions is treated as binary/categorical classifiers. The CNN-LSTM was trained using an L2 loss on $46,080$ generated examples, and tested on $11,520$ examples.
	
	\subsubsection{List of labels for pre-training: } For learning features specific to the room layout images that were are part of the Visual Hints setup, we pre-trained the visual part of the model on 42 different tasks. The following are the list of those 42 binary tasks:
	\{\textsc{death room y, death room n, literal, random number, room importance, colorway y,
		colorway n, living room, garden, driveway, bedroom, bathroom, corridor, shed,
		pantry, backyard, supermarket, kitchen, street, 0, 1, 2, 3, 4, 5, 6,
		7, 8, 9, 10, 11, player in y, player in n, room inaccessible=0,
		room inaccessible=1, room inaccessible=2, room inaccessible=3, rooms in colorway=0,
		rooms in colorway=1, rooms in colorway=2, rooms in colorway=3, rooms in colorway=4}\}. Pre-training on these multiple binary tasks improves the feature extraction capability of the CNN that is suited for tasks like death-room detection or `kitchen' extraction from text, which is ultimately going to be useful for the reinforcement learning agent.

	% We plot the F1 score function of the threshold for each class (see Figure~\ref{fig:simple_cnn_F1_score}).
	We plot the F1 score function of the threshold for each class in Figure~\ref{fig:simple_cnn_F1_score}. The F1 score at the threshold of $0.5$ is nearly perfect for all the easy questions (around or above $0.999$) and still reasonably good for hard questions (around $ 0.8 $). We observe that the most difficult task is to find the number of \textit{inaccessible rooms} for \textit{inaccessible rooms = 2} and \textit{inaccessible rooms = 3}. This result was expected, because it requires finding a relation between rooms with only the clue of open doors, compared to the other hard questions: \textit{number of rooms in the color path}, where all rooms are fully green. We also see that for these two questions, the F1 score curve is less flat, showing the difficulty that the network has in discriminating the answer in two groups.
	
		\begin{table*}[ht]
		\centering
		\begin{tabular}{|c|c|c|c|c|c|c|}
			\hline
			& \multicolumn{3}{c|}{LeDeepChef} & \multicolumn{3}{c|}{Visual-Lingual Model (our)} \\
			& \multicolumn{3}{c|}{} & \multicolumn{3}{c|}{} \\ \hline
			Success Rate & Train & Valid & Test & Train & Valid & Test \\ \hline
			Navigation (6 rooms) & 0.93 & 0.96 & 0.90 & 0.86 & 0.94 & 0.74 \\
			Navigation (9 rooms) & 0.78 & 0.83 & 0.56 & 0.62 & 0.47 & 0.30 \\
			Navigation (12 rooms) & 0.64 & 0.67 & 0.31 & 0.47 & 0.46 & 0.13 \\
			Navigation (All) & 0.79 & 0.82 & 0.59 & 0.65 & 0.68 & 0.39 \\
			Non-Navigation & 0.91 & 0.94 & 0.91 & 0.92 & 0.95 & 0.85 \\ \hline
			Total & 0.82 & 0.86 & 0.68 & 0.73 & 0.76 & 0.52 \\ \hline
		\end{tabular}
		\caption{Performance of our baseline model in terms of success rate compared to the state-of-the-art model, LeDeepChef~\cite{adolphs2019ledeepchef}.}
		\label{table:2}
	\end{table*}
	
	\subsection{Spatial Attention Analysis of the CNN output}
	
\begin{figure}[t]
		\centering
		\includegraphics[width=0.4\textwidth]{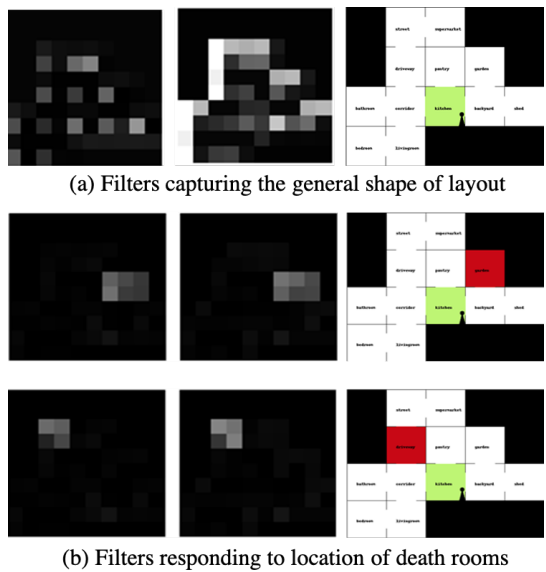}
		\caption{CNN outputs indicating the evolution of two maps based on variance in the the presence and position of the death room.}
		\label{fig:death_room_y_selectivity}
	\end{figure}

	To further analyze the features learned by our visual model, we try to understand the information that is encoded at the output of the CNN layer; and whether the LSTM layer makes the links between the different rooms as intended when we designed the CNN-LSTM architecture. A visual examination of the output of the CNN (Figure~\ref{fig:death_room_y_selectivity})  reveals that certain filters in the CNN extract the general shape and highlight individual rooms. In Figure~\ref{fig:death_room_y_selectivity}(b), we show some room layouts and their evolution when we vary the location of the death room on the map. Some convolutional filters are receptive to specific characteristics like the presence of a death room, a color path, a player, etc. -- this further illustrates that our visual model learns the semantics of the room layouts, which are useful features for the final RL task.
	
	\begin{table}[t]
		\centering
		
		\begin{tabular}{|c|c|c|c|}
			
			\hline
			\textbf{Number of games} & \textbf{Train} & \textbf{Valid} & \textbf{Test}   \\
			\hline
			Navigation (6 rooms) & 1041 & 52 & 124  \\
			\hline
			Navigation (9 rooms) & 1041 & 52 & 124  \\
			\hline
			Navigation (12 rooms) & 1041 & 52 & 124  \\
			\hline
			Navigation (All) & 3123 & 156 & 372   \\
			\hline
			Non-Navigation & 1317 & 66 & 142  \\
			\hline
			Total & 4440 & 222 & 514  \\
			\hline
			
		\end{tabular}
		
		\caption{Number of navigation based and non-navigation based games in the Cooking World.}
		\label{table:1}

	\end{table}

	\subsection{Results of Baseline Multimodal Agent}
	\label{subsec:results_baseline_mm_agent}
	
	Our multimodal agent was evaluated against the previous best method viz. LeDeepChef~\cite{adolphs2019ledeepchef} to compare the quality of generated action commands for solving the cooking games. The objective in these games is to gather all the ingredients needed for cooking a meal by following a recipe available in the kitchen. The agent may not start at the kitchen,  and has to navigate through the rooms to reach the kitchen. Upon reaching, the agent then has to follow the descriptions from the recipe to interact with the objects in the scene, and finally make a meal. Intermediate rewards are obtained along the way. 
	% to aid in the RL training.
	
	Based on the type and difficulty of the games, we divide them into two distinct categories. (i) \textsc{Navigation Games}: These are games that require some form of navigation actions to reach a destination where non-navigation commands are used to finish the goal. We further subdivide these games into $3$ categories based on difficulty. Each navigation game consists of either $6$, $9$, or $12$ rooms which determine the difficulty. (ii) \textsc{Non-navigation Games}: These are games that usually start at the kitchen and can be solved by non-navigation based commands only. These games are typically easier to solve than the former type. Table~\ref{table:1} shows the number of games belonging to each of these categories.
	
	For our baseline model, we use the architecture setup as shown in Figure~\ref{fig:overall}; this simply concatenates the features from the visual and textual components of the observation. In our setup, we use the relatively simpler setting of {\em Clue First Room} as true, which means that the hint is always available in the room where the agent spawns. Therefore, for navigation-based games, the goal of the agent is to first reach the kitchen and then complete the task by using non-navigation commands. Since the multimodal policy would be beneficial in the navigation part of the game, we use a flag that keeps track of whether the kitchen has been reached. Thus, for navigation games, the part of the game from the spawning location to the kitchen is handled by the visual-lingual policy; whereas the non-navigational commands in the kitchen are handled by the text-based policy only.
	
	% \textcolor{red}{Please argue why not beating SOTA is Ok and how this is just a baseline method}
	
	Table~\ref{table:1} shows the performance of our agent in terms of success rate as compared to LeDeepChef~\cite{adolphs2019ledeepchef}. Although our simple multimodal policy does not outperform the state-of-the-art method, our baseline model provides an evaluation method based on navigation and non-navigation games; and the performance metrics obtained by simple features' fusion-based policy models. However, the primary purpose of this evaluation is not to beat the SOTA technique; but rather to show the value in utilizing the visual information in addition to the textual information in a given game. Furthermore, Table~\ref{table:1} is intended to be incomplete -- this work is a call to other contenders in the field of TBGs and RL to utilize the multiple modalities of vision and text to beat the reigning SOTA on this problem. There is a large possibility of improved performance with algorithmic improvements and novelty. 
	
	Furthermore, the problem settings that we have provided in Section~\ref{subsec:generating_hints_from_tw} enable the design of more challenging SOTA benchmarks. The most important contribution of this work lies in providing the extremely customizable \vizhints\ environment, where simple control variable alterations can produce problems that differ significantly in terms of their challenge to existing RL techniques. Furthermore, the generation of the visual hints is completely automated, and agnostic of the specific game or even domain that the agent finds itself in. We hope this spurs more research on the \vizhints\ environment, and the creation of diverse algorithmic techniques that handle different variants of this complex and challenging environment.
	
	% There is a large possibility of improving the performance by algorithmic improvements. Furthermore, various new problem settings can be tried on our environment using simple control variable alterations.

	\section{Conclusion}
	\label{sec:conclusion}
	
	To bridge the gap between purely unimodal text or vision based approaches in reinforcement learning, we present a novel environment called \vizhints that uses both visual and textual information for multimodal reinforcement learning. We use the cooking environment in TextWorld as a base upon which we add visual information in the form of room layouts as hints located at various rooms. Our proposed environment showcases a high level of detailing, allowing multiple RL problem settings to be emulated seamlessly by changing a few options. The final goal is to learn RL agents that use such multimodal information efficiently to outperform their unimodal counterparts. We present a baseline multimodal agent using CNN based feature extraction from visual hints and LSTM for textual feature extraction. Furthermore, we perform pre-training on the visual model using a $42$-task classification problem for predicting semantic information about the game world. 
	
	We achieve a high F1-score on most labels, while outputs related to predicting relations between the rooms prove to be the most difficult. We also train a baseline visual-lingual policy network that concatenates visual and textual features to perform model-free reinforcement learning to set up a baseline score. Our proposed environment poses numerous technical challenges like partial observability, shared feature learning, and large action space; all of which can independently be spotlighted using various settings in the environment. We believe that our proposed \vizhints\ environment and accompanying dataset will provide a common framework to measure algorithmic progress in the multimodal RL community.
	
	\bibliographystyle{aaai}
	\bibliography{twmap}

\end{document}